\documentclass[letterpaper, 10 pt, conference]{ieeeconf}
\IEEEoverridecommandlockouts
\usepackage[T1]{fontenc}
           
\usepackage{booktabs}
\usepackage{multirow} 
\usepackage{makecell}
\usepackage{xcolor}
\usepackage{xspace}
\usepackage{colortbl}
\usepackage[table]{xcolor}

\usepackage{balance}
\usepackage{silence}

\usepackage{capt-of}
\usepackage[caption=false,font=normalsize,labelfont=sf,textfont=sf]{subfig}
\usepackage{textcomp}
\usepackage{stfloats}

\IEEEoverridecommandlockouts

\usepackage{graphicx}
\usepackage{picinpar}
\usepackage{url}
\usepackage{flushend}
\usepackage[latin1]{inputenc}
\usepackage{soul}
\usepackage{pifont}
\usepackage{alltt}
\usepackage{enumerate}
\usepackage{siunitx}
\usepackage{pbox}
\usepackage{amsmath,amssymb,amsfonts}
\usepackage{mathrsfs}
\usepackage[mathscr]{eucal}
\usepackage{bm}
\usepackage{graphbox}
\usepackage[export]{adjustbox}

\usepackage[ruled,vlined]{algorithm2e}

\definecolor{LightCyan}{rgb}{0.8,0.8,1.0}
\definecolor{LightRed}{rgb}{1.0,0.8,0.8}
\definecolor{LightGreen}{rgb}{0.8,1.0,0.8}
\definecolor{LightYellow}{rgb}{1.0,1.0,0.8}

\ActivateWarningFilters[balance]

\makeatletter
\let\NAT@parse\undefined

\makeatother

\usepackage{cite}

\usepackage{hyperref} 
\hypersetup{
    hidelinks,
    colorlinks=true,        
    linkcolor=black,
    citecolor=black,
    filecolor=black,
    urlcolor=blue
}

\def\sysname{\texttt{SkySearcher}\xspace}
\title{\fontsize{23pt}{27pt}\selectfont 
SkySearcher: High-Altitude Onboard VLM Reasoning for Aerial Semantic Search
}

\author{
\thanks{Disclosure of AI usage: Anthropic-Claude models were used for assisting in code development of \sysname.}
}

\author{
  \begin{tabular}{c}
    Giacomo Vianello\textsuperscript{1} \qquad
    David Morilla-Cabello\textsuperscript{2} \qquad
    Juan Dendarieta Serrano\textsuperscript{2}
  \end{tabular}
  \\[1ex]
  \begin{tabular}{c}
    Ruggero Carli\textsuperscript{1} \qquad
    Eduardo Montijano\textsuperscript{2}
  \end{tabular}
  \thanks{
    This work was partially funded by project T45\_23R, by grants AIA2025-163563-C31, PID2024-159284NB-I00, funded by MCIN/AEI/10.13039/501100011033 and ERDF, the Office of Naval Research Global grant N62909-24-1-2081
    \newline
    \textsuperscript{1}Department of Information Engineering (DEI) at the University of Padova, Italy. \{giacomo.vianello.2@phd, carlirug@dei\}.unipd.it.
    \newline
    \textsuperscript{2}Instituto de Investigacion en Ingenieria de Aragon, Universidad de Zaragoza, Spain. \{davidmc, jdendarieta, emonti\}@unizar.es
  }
}

\begin{document}
\maketitle

\begin{abstract}
Recent advances in Vision-Language Models (VLMs) have improved semantic and visual reasoning, enabling UAVs to search for open-vocabulary queries in unstructured environments at high altitude. However, existing systems mainly follow low-altitude strategies, focusing on close-range iterative exploration, requiring storage-intensive dense 3D metric-semantic reconstructions or relying on large cloud-serviced VLMs assuming constant internet access. We propose \sysname, an aerial semantic search navigation method designed for large outdoor scenes. \sysname integrates onboard VLM reasoning with a lightweight probabilistic representation that guides an information-driven exploration. The VLM extracts semantic information at high altitude about the target or cues informing about its location. This information is projected to the ground using the camera frustum instead of limited-depth sensors, actively exploiting the expansive field of view offered by elevated sensing. The information regarding the presence or absence of semantic cues is then integrated into a probabilistic map by modeling the spatial belief of the query, maintaining a task-driven memory for the mission and guiding the exploration towards semantically promising regions. Extensive simulations for Aerial Object Goal Navigation on public benchmarks demonstrate that \sysname significantly outperforms state-of-the-art baselines in success rate, flight time, and navigation precision, while a deployment of \sysname running fully onboard a drone demonstrates the applicability of our approach.

\noindent
\textit{Code and dataset will be publicly released upon acceptance and can be reviewed \href{https://anonymous.4open.science/r/SkySearcher-AF10/}{here}.}
\end{abstract}

\section{Introduction}\label{I}

Aerial platforms operating at high altitude benefit from a wide effective field of view, which has proven valuable for tasks such as aerial surveillance~\cite{sun2024moving}, post-disaster search and rescue~\cite{abdellatif2025pdsr}, and facility inspection~\cite{phan2024deep}. Recent Vision--Language Models (VLMs) enable extracting rich semantic information and reasoning about the environment, which is promising for interpreting this broad spatial context and perform tasks autonomously.

\begin{figure}[t]
    \centering
    \includegraphics[width=0.98\linewidth]{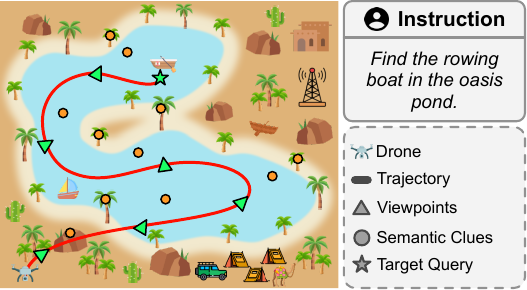}
    \vspace{-7pt}
    \caption{Illustrative \sysname mission in a large-scale outdoor environment. Given an open-vocabulary instruction, the drone autonomously explores the unknown environment and reaches the specified target by exploiting intermediate semantic cues (e.g., rocks, palm trees, or lake) encountered en route.}
    \label{fig:teaser}
\end{figure}

The capabilities of VLMs have had a great impact in semantic tasks such as Aerial Object Goal Navigation (Aerial ObjectNav)~\cite{batra2020objectnav}. In this setting, an autonomous aerial agent explores an unknown environment to identify an open-vocabulary target queried by the user while acquiring information from its sensors, as depicted in Fig.~\ref{fig:teaser}. However, existing solutions do not leverage the high-altitude flying capabilities of drones explicitly and usually adopt low-altitude search strategies. These strategies aim at exploration using range sensors and object detectors, and sometimes require dense reconstructions of the environment, such as semantic voxel maps~\cite{alama2025rayfronts, kim2025raven}, and complex trajectory planning for efficient navigation\cite{chen2026airhunt, laina2025findanything}. On the other hand, methods leveraging VLMs usually assume the availability of large cloud-served models to reason on images during the search mission \cite{chen2025codrone, ji2025towards}, restricting the applicability to settings with stable network connectivity and incurring on constant API usage costs. Furthermore, other methods give the VLM the full mission control by tasking it to directly analyze images, infer distances, and output movement commands \cite{xiao2025uav, hu2025see}. However, these models have not been trained for these tasks and can produce unpredictable behaviors.

In this work, we address the aforementioned limitations by introducing \sysname, a method that leverages the visual perception and semantic reasoning capabilities of a small VLM to enable fully onboard, open-vocabulary aerial ObjectNav at high altitude. Within the proposed system, the VLM processes camera observations to identify both the query and a set of semantically informative cues that may inform about its location. These observations are projected on a lightweight map representation integrating geometric information about already explored regions and semantic information expressing the spatial belief of the query. This representation supports the formulation of an information-theoretic optimization problem for selecting the next-best viewpoint for the high-altitude UAV exploration.

To validate our approach, we perform extensive Aerial ObjectNav experiments in a photorealistic simulator over a variety of objects and large environments. Results show that \sysname improves success rate by over 58\% the strongest baseline while requiring no cloud connectivity or dense metric-semantic reconstructions. To further demonstrate the practical feasibility, we deploy \sysname running fully onboard a quadrotor in a real-world scenario.

To summarize, the contributions of \sysname are:
\begin{itemize}
    \item An aerial semantic search strategy from high altitude that relies on local VLM reasoning over images and frustum projection, exploiting the extensive visibility of high vantage viewpoints while running fully onboard.
    \item A lightweight probabilistic representation that fuses metric and semantic information into a single map, encoding the spatial belief of the query from intermediate semantic cues without dense reconstructions.
    \item An information-based next-best-viewpoint optimization for planning in open-vocabulary semantic search.
    \item An open-source implementation of the full system in ROS2, deployable both in photorealistic simulation and on real hardware, supporting reproducibility and reuse by the community.
\end{itemize}
\section{Related Work}
\label{sec:related_work}

\subsection{Object Goal Navigation} 
Introduced in \cite{anderson2018evaluation} and \cite{batra2020objectnav}, Object Goal Navigation is a semantically grounded task that requires an autonomous agent to navigate to an instance of a specified object category within an unknown environment, relying strictly on onboard observations without prior maps. Some works previously tackled this task via Reinforcement Learning for closed-set categories \cite{gireesh2022object}, but recent advances in the field of visual and semantic reasoning proved that the same task can be carried out by leveraging foundation models. For example, VLFM \cite{yokoyama2024vlfm} uses VLMs to score exploration frontiers on value maps, FindAnything \cite{laina2025findanything} grounds open-vocabulary queries into 3D scene representations for mapless zero-shot navigation, and ApexNav \cite{zhang2025apexnav} balances semantic reasoning and geometric exploration via cue strength.
However, prior works mainly target small, semantically uniform indoor scenes, where structured layouts and consistent semantic hierarchies aid object localization. In contrast, we address the aerial domain in large-scale outdoor environments (e.g., cities, waterfronts), characterized by expansive search spaces, sparse targets in unstructured terrain, and weaker semantic priors to guide exploration.

\subsection{Aerial ObjectNav} 
Interest in the ObjectNav task has recently expanded toward aerial applications, driven by the emergence of powerful VLMs capable of open-vocabulary semantic reasoning. AirHunt~\cite{chen2026airhunt} decouples slow semantic reasoning from fast planning through a shared value map to avoid stop-and-infer flight, while See, Point, Fly~\cite{hu2025see} skips 3D reconstruction entirely and predicts waypoints as 2D points directly on the image. SkyVLN~\cite{li2025skyvln} enforces NMPC-safe navigation for urban flight, PRPSearcher \cite{ji2025towards} leverages multi-modal large language models to mimic human three-tier cognition by building specialized maps, and LMPath~\cite{diller2026lmpath} moves semantic reasoning outside the flight loop altogether, using an LLM to infer likely environmental contexts for a target and a vision foundation model to extract corresponding regions from satellite imagery, which are then used to generate search paths. All of these, however, rely on large, cloud-hosted VLMs for semantic reasoning, incurring the latency, cost, and connectivity dependence this entails. In contrast, \sysname pairs a small-scale, fully onboard VLM with a lightweight semantic representation, enabling low-latency, open-vocabulary object search directly on the flight computer, with no recurring API costs and no reliance on network connectivity.

\subsection{Semantic Mapping}
To aid exploration, it is customary to maintain spatial memory of what has already been observed and where the target is likely to be, though existing approaches differ substantially in how much of this memory they retain online. RayFronts~\cite{alama2025rayfronts} fuses near-range semantic voxels with far-range rays to prune the search space; RAVEN~\cite{kim2025raven} extends this representation with a behavior tree for dual-horizon planning. HALO~\cite{tao2026halo} instead maintains a dense metric-semantic reconstruction for high-altitude monocular exploration, trading a substantially richer mapping pipeline for more detailed geometric and semantic reasoning. Moving toward lighter representations, Memory over Maps~\cite{zhou2026memory} avoids global 3D reconstruction altogether, retrieving and re-ranking posed RGB-D keyframes to build a sparse, query-specific estimate on demand, while WildOS~\cite{shah2026wildos} sits at the memory-light end of the spectrum, guiding legged exploration with a sparse navigation graph paired with a vision foundation model that evaluates traversability, visual frontiers, and object similarity.

This spectrum reflects a consistent trade-off: denser representations afford richer geometric and semantic reasoning at a proportional memory and computational cost, while lighter alternatives reduce this overhead at the expense of a less explicit global representation of the explored space. \sysname instead maintains a lightweight map representation---a 2D geometric coverage and a probabilistic semantic map---built and updated online from onboard observations alone, providing the spatial memory required for long-horizon outdoor search without a dense volumetric reconstruction.
\section{Problem and Solution Overview}
\label{sec:method}

\begin{figure*}[t]
    \centering
    \includegraphics[width=\linewidth]{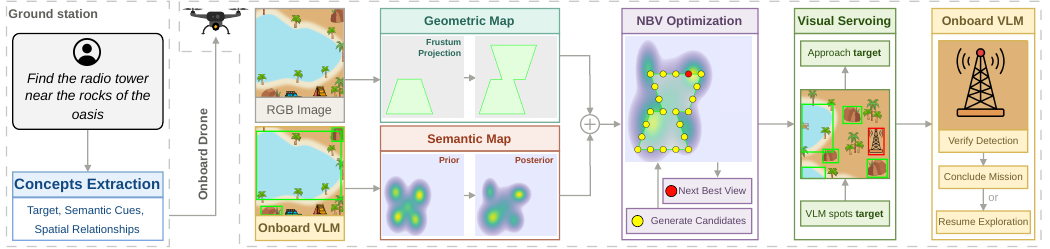}
    \vspace{-18pt}
    \caption{\textbf{System Overview}. A ground-station LLM parses the user query $Q$ into the target, semantic cues, and their expected co-occurrence distance, and variability. Onboard, the drone maintains a \textbf{geometric coverage map} and a \textbf{probabilistic semantic map}, while a compact VLM detects and verifies the target and cues in the RGB stream. \textbf{NBV optimization} fuses this geometric and semantic evidence to select the next viewpoint, until \textbf{visual servoing} approaches the target detection and confirms (or rejects) it.}
    \label{fig:method}
    \vspace{-18pt}
\end{figure*}

\subsection{Problem Definition}

Given an aerial robot with a known pose obtained via GPS or localization methods, Aerial ObjectNav aims at autonomously finding the spatial location of a semantic query in a bounded and unknown environment. In this paper, we focus our attention on large-scale outdoor environments.
The semantic query, denoted as $Q,$ is provided by the user via natural language (e.g.,  \textit{``find the rowing boat in the lake''}). Our objective is to infer the spatial location of the query. 
In this setting, a mission is considered successful if the agent arrives within a specified distance of the target location and confirms mission termination.  

\subsection{\sysname overview}

An overview of how our solution works is illustrated in Fig. \ref{fig:method}.
\sysname operates the drone at an arbitrary fixed flight altitude with a single onboard camera rigidly mounted at a fixed downward pitch.
At mission start, we extract a list of semantic concepts from the query, including a primary target object, a set of related semantic cues, and their spatial relationships. This is done in the ground station by using a cloud LLM, and it is the only step in our pipeline that relies on external connectivity; all subsequent reasoning is performed onboard for the remainder of the mission.

As the drone explores, it takes images at \textit{viewpoints} and combines them in 
a lightweight map representation of the environment and its relationship to the task (Sec.~\ref{sec:mapping}): a \textit{geometric coverage map}, built by incrementally projecting the onboard camera frustum onto the ground plane, tracks explored regions, while a \textit{probabilistic semantic map} maintains a belief over the presumed target location, updated via Bayesian filtering using the predictions given by the compact onboard VLM.

The map drives a high-level planner (Sec.~\ref{sec:planner}) that solves an information-theoretic next-best-view problem to select the drone's next \textit{viewpoint}. Once the target is detected with sufficient confidence, a visual servoing module takes care of a final verification to prevent false positives caused by the high-altitude vision.

\section{Lightweight Map Representation}
\label{sec:mapping}

To keep the computational burden of \sysname bounded, we exploit the fact that the drone is flying at a high altitude to consider a simplified 2D representation of the search space.
We define this space as a bounded ground domain $G \subset \mathbb{R}^2$, and denote by $x \in G$ a candidate ground location where the target may lie.
We denote by $w_t = (x_t, y_t, \psi_t)$ the \textit{viewpoint} where the drone acquires an image at time $t$, i.e., its position and yaw heading at the fixed operating altitude.

We use the measurements taken at these viewpoints to compute two representations.
The geometric coverage map, $\mathcal{C}_t \subseteq G,$ maintains a persistent record of previously observed ground regions, distinguishing explored from unexplored areas of the environment.
The probabilistic semantic map is modeled via a Bayesian belief, $\pi_t(x)$, representing our confidence that the semantic query $Q$ holds true at each location $x \in G$. 
Formally, given the query $Q$, the semantic map estimates
\begin{equation}
    \label{eq:posterior_def}
    \kern -5pt \pi_t(x) := P\big(Q_x \mid Z_{0:t}\big),\; \int_{G} \pi_t(x) \, dx= 1,\; \pi_t(x) \geq 0,
\end{equation}
where $Z_{0:t}$ denotes the history of observations acquired from mission start up to time $t$, and $Q_x$ denotes the event that the query $Q$ is satisfied at $x$.
Additionally, we store the list of past viewpoints, $\mathcal{W}=\{w_0,\ldots,w_t\},$ to favor novel viewpoints during planning.

\subsection{Geometric Coverage Map Update}
\label{sec:geometric_coverage}

At each viewpoint, $w_t$, where an image is acquired, using the camera calibration parameters we intersect the camera frustum with the ground to obtain the visible ground footprint, $\mathcal{F}_t\equiv\mathcal{F}(w_t)$.
We fuse the footprints observed at each step recursively, keeping track of the full extent of the environment explored so far (Fig.~\ref{fig:method}).
In practice, this update is carried out via morphological closing with buffer, $\epsilon_{\text{buf}}$, to eliminate numerical gaps between consecutive footprints,
\begin{equation}
    \mathcal{C}_t = \Big[(\mathcal{C}_{t-1} \oplus \epsilon_{\text{buf}}) \cup (\mathcal{F}_t \oplus \epsilon_{\text{buf}})\Big] \ominus \epsilon_{\text{buf}}, \qquad \mathcal{C}_0 = \emptyset,
    \label{eq:coverage_update}
\end{equation}
where $\oplus$ and $\ominus$ denote the Minkowski sum and difference~\cite{soille2003morphological}, respectively.

\subsection{Probabilistic Semantic Map Update}
\label{sec:prob_map}

We update $\pi_t(x)$ using recursive Bayes, under the standard assumptions of the query not changing over time and conditionally independent observations,
\begin{equation}
    \label{eq:bayes_normal}
    \pi_t(x) = \eta \cdot \pi_{t-1}(x) \cdot P(Z_t \mid Q_x),
\end{equation}
where $\eta$ is a normalization constant and $P(Z_t\mid Q_x)$ is the likelihood of the observation acquired at time $t$. 
The central question is therefore how to model the semantic observations and their likelihood using the onboard VLM. 
In the following, we propose a procedure to approximate these quantities for our problem.

\subsubsection{VLM Observations}
\label{sec:VLMobs}
We exploit the ability that VLMs have to detect open vocabulary objects in images to develop our whole semantic perception pipeline.
Prior to the mission, we extract from the query $Q$ the concepts of a primary target, denoted by $C_\mathrm{target}$, a list of associated semantic cues $C = \{c_i\}_{i=1}^n,$ and the spatial relationship expected between each cue and the target, captured by an expected distance $d_c$ and variance $\sigma_c$, $c=1,\ldots,n$; this decomposition is performed once, offline, via a cloud-hosted LLM.
For instance, for $Q = $ \textit{``find the rowing boat in the lake''}, one may expect the primary target to be $C_\mathrm{target} = \text{boat},$ the cues $C = \{\text{lake}, \text{shore}, \text{dock}\}$, and for the lake $d_c \approx 0$ and small $\sigma_c$, describing that the boat should be somewhere inside the lake.

Then, during the mission, we query the local onboard VLM to provide detections of $C_\mathrm{target}$ and $C$ at each viewpoint.
The VLM returns a bounding box and a confidence score for each detection, which are the measurements we use to update $\pi_t(x)$.

\subsubsection{Measurement Likelihood}

Each cue detected by the VLM from $w_t$ provides a positive evidence. 
Additionally, when the primary target is \emph{not} detected, we introduce a dedicated negative-evidence term to penalize the currently observed region. 
Making the simplifying assumption of conditional independence among the cues, we can treat their likelihoods separately and multiply them
\begin{equation}
    \kern -5pt P(Z_t\mid Q_x) \simeq P(\neg Z_t^{target} \mid Q_x) \prod_{c=1}^{n} P(Z_t^c \mid Q_x).
    \label{eq:total_likelihood}
\end{equation}

\begin{figure}[t]
    \centering
    \includegraphics[width=\linewidth]{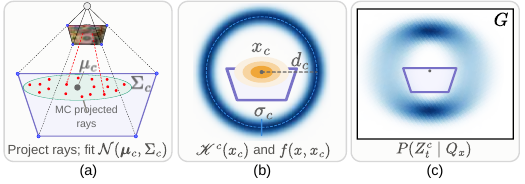}
    \vspace{-23pt}
    \caption{Construction of the per-concept likelihood: (a) VLM detections are projected to the ground plane and fitted to $\mathcal{N}(\boldsymbol{\mu}_c,\Sigma_c)$; (b) combining the resulting kernel $\mathscr{K}^c(x_c)$ with the spatial term $f(x,x_c)$ yields the concept likelihood $P\big(Z_t^c \mid Q_x\big)$, shown in (c).}
    \vspace{-10pt}
    \label{fig:mapping}
\end{figure}

\textit{Positive Evidence.}
Our model combines two factors: the confidence of the VLM detection of $c$ at $x_c\in G$, denoted $\mathscr{K}^c(x_c),$
and the plausibility of the spatial relationship between $c$ located at $x_c$ together with the primary target at $x$, denoted by $f(x,x_c)$,
\begin{equation}
    P\big(Z_t^c \mid Q_x\big) = \int_{x_c\in G} \alpha(x_c) \cdot \mathscr{K}^c(x_c) \cdot f(x,x_c) \cdot dx_c,
    \label{eq:concept_likelihood}
\end{equation}
where the integral marginalizes over all ground positions $x_c$ at which cue $c$ may be observed, and $\alpha(x_c)$ is a redundancy discount that down-weights detections of concepts already observed multiple times at $x_c$, e.g., a repeatedly-observed lake contributes less to subsequent updates, Fig.~\ref{fig:mapping} (c).

We model $f(x,x_c)$ as a Gaussian centered at expected distance given prior to the mission, $d_c$, with variance $\sigma_c$ (Sec.~\ref{sec:VLMobs}), Fig.~\ref{fig:mapping} (b).
The confidence term $\mathscr{K}^c(x_c)$ translates the bounding box provided by the VLM to spatial locations in $G$.
To do this, we project it onto the ground plane by casting rays from Monte Carlo-sampled interior points, discarding those that fall outside $G$, and fit the resulting ground point cloud to a Gaussian $\mathcal{N}(\boldsymbol{\mu}_c, \Sigma_c)$ via sample mean and covariance, Fig.~\ref{fig:mapping} (a). We capture the resulting geometric uncertainty with a \textit{Flat-Top Kernel} centered at $\boldsymbol{\mu}_c$,
\begin{equation}
    \label{eq:flat_top_kernel}
    \mathscr{K}^c(x_c) = \gamma_c \cdot \exp\left(-\frac{d_E(x_c)^2}{2\nu^2}\right),
\end{equation}
where $\gamma_c$ is the VLM detection confidence of the object and $\nu$ a decay factor (in meters). The term $d_E(x_c)$ is the Euclidean distance from the boundary of the ellipse, defined as zero wherever the Mahalanobis distance $D_M(x_c) \leq 1$ (via $\Sigma_c$), i.e., inside the ellipse, and as the actual boundary distance otherwise. This yields uniform confidence $\gamma_c$ throughout the detection ellipse, decaying only beyond its boundary, and avoids the artificial confidence peak at $\boldsymbol{\mu}_c$ that a plain Gaussian would place at a single, arbitrarily-chosen point, Fig.~\ref{fig:mapping} (b). 

\textit{Negative Evidence.}

When cues $c \in C$ are detected at viewpoint $w_t$, the previous term increases the belief of the query at distances around $d_c,$ regardless of whether the target itself lies within the currently observed region $\mathcal{F}_t$. 
If $C_\mathrm{target}$ is not visible in $\mathcal{F}_t$, then this likelihood must be penalized. 
We thus define the non-observation likelihood, parameterized by $\beta \in [0, 1)$, as
\begin{equation}
    P(\neg Z_t^{target} \mid Q_x) =
    \begin{cases}
    1 - \beta & \text{if } x \in \mathcal{F}_t \text{ and }\\
    & \hskip 0.4cm  C_{\mathrm{target}} \text{ is absent}\\
    1.0 & \text{otherwise}.
    \end{cases}
\end{equation}
This construction ensures two properties: (i) belief propagated by contextual cues into the currently observed region $\mathcal{F}_t$ is immediately down-weighted by $(1-\beta)$, preventing spurious mass from persisting in already-explored terrain; and (ii) for unobserved cells $x \notin \mathcal{F}_t$, the likelihood is not penalized, favoring exploration toward yet-unvisited candidate locations.

\section{High-Level Planning and Exploration Candidates}
\label{sec:planner}
To determine where the drone should move next, we introduce a planning strategy built on the lightweight map representation of Sec.~\ref{sec:mapping}. 
It operates in two phases: (i) an information-based next-best-view (NBV) planner selects, at each step, the most informative viewpoint from a set of candidates (Sec.~\ref{sec:nbv}); and (ii) once the VLM detects the primary target, a visual-servoing phase closes the distance to it while repeatedly querying the VLM for verification, counteracting false positive detections (Sec.~\ref{subsec:visual_servoing}).

\subsection{Information-based Next-best-viewpoint Optimization}
\label{sec:nbv}
Our planning approach guides the drone to explore the environment while actively gathering semantic evidence related to $Q$. We formulate this objective as a NBV optimization problem:
\begin{equation}
    w^\star = \arg\max_{w \in \mathcal{W}_t} J(w).
\end{equation}
Here, $\mathcal{W}_t = \{w_i= (x_i, y_i, \psi_i)\}$ is a set of candidate waypoints evaluated at step $t$.
Our planning pipeline addresses this optimization through two core modules: (i) a candidate generation mechanism, and (ii) an information-theoretic objective function $J(\cdot)$.

\subsubsection{Waypoint Generation}
\begin{figure}[t]
    \centering
    \vspace{-12pt}
    \includegraphics[width=\linewidth]{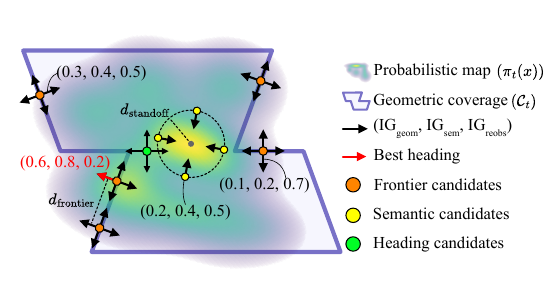}
    \vspace{-30pt}
    \caption{Example of NBV planning. Candidate viewpoints are sampled based on their perspective interest, and each is evaluated using three criteria that consider geometric gain, semantic gain, and reobservation gain. The planner then selects the candidate that maximizes this overall reward.}
    \label{fig:planning}
\end{figure}

At each planning step $t$, the candidate set $\mathcal{W}_t$ is formed by combining three complementary types of viewpoints (Fig.~\ref{fig:planning}). \textit{Frontier candidates} are uniformly sampled along the boundary $\partial \mathcal{C}_t$, equispaced by $d_{\text{frontier}}$ (in meters). At each position, four orientations are assigned---aligned with the inward and outward normals and both boundary tangents---systematically driving the drone to expand the coverage. \textit{Semantic candidates} are placed around the spatial centroids $\boldsymbol{\mu}_c$ of detected cues, clustered beforehand to eliminate duplicates. For each cluster, four orthogonal viewpoints are positioned at a standoff distance $d_{\text{standoff}}$, facing $\boldsymbol{\mu}_c$, allowing the drone to inspect promising cues from multiple angles. Finally, \textit{heading diversity candidates} consist of in-place yaw rotations evaluated at the current drone location, to inspect previously unobserved viewing directions, at negligible travel cost.

\subsubsection{Objective Function}
We evaluate each element $w \in \mathcal{W}_t$ with an objective function that trades off the informativeness of a waypoint against the cost of reaching it:
\begin{equation}
    \label{eq:objective_planner}
    J(w) = \text{IG}(w) - \gamma \| \boldsymbol{p}_w - \boldsymbol{p} \|_2,
\end{equation}
where $\boldsymbol{p}_w, \boldsymbol{p} \in \mathbb{R}^2$ denote the positions of the candidate viewpoint and the drone, respectively, and $\gamma > 0$ weights the distance penalty. The total information gain $\text{IG}(w)$ combines geometric exploration, semantic exploitation, and re-observation terms,
\begin{equation}
    \text{IG}(w) = \text{IG}_{\text{geom}}(w) + \lambda_{\text{sem}} \text{IG}_{\text{sem}}(w) + \lambda_{\text{reobs}} \text{IG}_{\text{reobs}}(w),
    \label{eq:InfGain}
\end{equation}
where $\lambda_{\text{sem}}, \lambda_{\text{reobs}} \in \mathbb{R}$ are free, tunable weights. We describe in detail each term of Eq. \eqref{eq:InfGain} below.

\textit{Geometric Information Gain}, $\text{IG}_{\text{geom}}(w)$, rewards pure geometric exploration, measuring the unobserved ground area the candidate viewpoint covers, 
\begin{equation}
    \text{IG}_{\text{geom}}(w) = \int_{x\in\mathcal{F}(w) \setminus \mathcal{C}_t} \pi_0 \, dx,
\end{equation}
where $\pi_0$ is the uniform prior probability with which \eqref{eq:bayes_normal} is instantiated, accumulating uniform mass over the newly-covered region, independent of semantic content.

\textit{Semantic Information Gain}, $\text{IG}_{\text{sem}}(w)$, blends exploration and exploitation, rewarding candidates that overlap with regions the semantic map already deems more likely than the uniform prior to contain the target,
\begin{equation}
    \text{IG}_{\text{sem}}(w) = \int_{x\in\mathcal{F}(w)} \max\big(\pi_t(x) - \pi_0, \, 0\big) \, dx.
\end{equation}
The $\max(\cdot, 0)$ clips cells whose belief has dropped below $\pi_0$ after repeated non-detections, preventing ruled-out regions from being rewarded.

\textit{Re-observation Information Gain}, $\text{IG}_{\text{reobs}}(w)$, rewards revisiting high-belief regions from novel perspectives. This is not only useful for recovering visibility lost to occlusions---comparatively rare at high altitude---but more generally for disambiguating borderline VLM detections through repeated observation. Reusing the recorded viewpoint history, $\mathcal{W}$, stored in the map, we define the angular novelty of observing $x$ from a candidate viewpoint $w_i$ with heading $\psi_i$ as
\begin{equation}
    g(x, w_i) = \frac{1}{2}\left(1 - \cos\!\Big(\min_{w \in \mathcal{W}:\, x \in \mathcal{F}(w)} |\psi - \psi_i|\Big)\right),
\end{equation}
i.e., the deviation of $\psi_i$ from the heading of the closest past viewpoint that already observed $x$. The angular novelty $g$ is maximal ($g\approx1$) for near-opposing views and vanishes for repeated ones. The gain then rewards angularly-novel views of already-explored, above-uniform-belief cells:
\begin{equation}
    \text{IG}_{\text{reobs}}(w) = \int_{x\in\mathcal{F}(w) \cap \mathcal{C}_t} \kern-25pt \max\left(\pi_t(x)- \pi_0, \, 0\right) g(x, w) \, dx.
\end{equation}
By explicitly rewarding angular diversity in high-probability regions, $\text{IG}_{\text{reobs}}(w)$ incentivizes the drone to re-examine candidate target locations from complementary viewing directions, uncovering occluded terrain behind trees and buildings while maintaining computational efficiency.

\subsection{Visual Servoing and False Positive Recovery}
\label{subsec:visual_servoing}

Upon detection of the primary target, the drone transitions from global exploration to a target-verification phase. Starting from the initial VLM detection, the drone iteratively approaches the target. At each step, the VLM confirms the target is still present, the drone corrects its bearing to recenter the target's bounding box in the image plane, and advances along the camera-to-object ray by a fixed step. This loop continues until either a depth-based standoff distance is reached or the VLM flags the detection as a false positive.

Upon reaching the standoff distance, a precise query to the local VLM performs a final verification. If verified, the mission ends successfully. If verification fails, or the target is lost during approach, the target location is added to a spatial blacklist and the drone returns to its pre-descent pose. Subsequent candidate generation then suppresses viewpoints within an exclusion radius of any blacklisted location, preventing repeated false-positive loops.
\section{Experiments \& Results}

\begin{figure}[t]
    \centering
    \includegraphics[width=\linewidth]{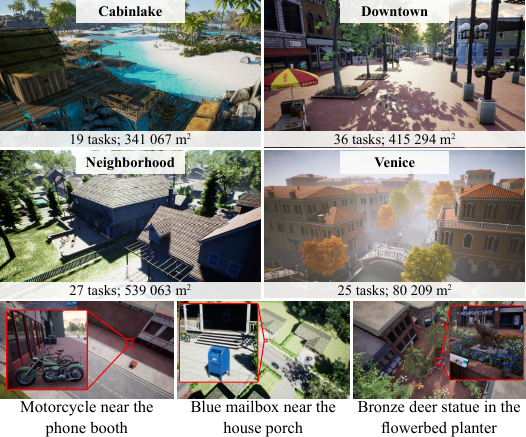}
    \vspace{-20pt}
    \caption{Overview of four simulation environments, labeled with the task count and the approximate spatial scale of each scene, along with representative examples of target queries used for evaluation.}
    \label{fig:environments}
\end{figure}

\noindent \textbf{Simulation Environments}.
We evaluate our system for Aerial ObjectNav tasks on Airsim~\cite{shah2017airsim}, a photorealistic drone simulator built on Unreal Engine 4.27. We follow the benchmark proposed by UAV-ON~\cite{xiao2025uav} using the same environments and target queries, extended to incorporate contextual information that guides the search. The benchmark includes four diverse, large-scale environments with 107 target query episodes depicted in Fig.~\ref{fig:environments}. Every benchmark episode is constrained to a maximum horizon of 50 decision steps, which are the viewpoints suggested by our planner.

\noindent \textbf{Implementation Details}. 
The simulated drone is equipped with an RGB-D camera pitched at \(-60^\circ\), a resolution of \(736 \times 736\) px with a square field of view of \(80^\circ\). We use its depth camera, clipped at 10\si{\meter}, and a resolution of \(256 \times 256\) px for the target approach. We use a LiDAR sensor with 20 m range for obstacle avoidance. We adapt the ROS2 navigation stack~\cite{macenski2020marathon2} for path planning and local navigation. The cruise speed of the drone is \(v_{m} = 2.0\) m/s, and the nominal flight altitude is \(z = 15\) m, with a minimum altitude during the visual servoing phase of \(z_\mathrm{min} = 3\) m.

We use Qwen3-VL-4B~\cite{bai2025qwen3} as our main VLM during the mission for detection and reasoning. This relatively small model can run onboard consumer robotic hardware that we target for our deployment. Benchmark episodes are executed on a desktop AMD Ryzen 9 9950X3D processor, an NVIDIA GeForce RTX 5070 Ti GPU with 16 GB of VRAM, and 32 GB of system RAM.

For the planner, we set the travel-cost penalty to \(\gamma = 6 \times 10^{-5}\), the semantic-importance weight to \(\lambda_\mathrm{sem} = 15.0\), and the re-observation reward to \(\lambda_\mathrm{reobs} = 10^{-3}\). For the Bayes filter and mapping components, we use $\alpha = \beta = 0.1$, a kernel decay of $\nu = 5$ m, and a coverage buffer of $\epsilon_\mathrm{buf} = 1$ m. For evaluation, both SR and OSR apply a goal-distance threshold of \(5\) m.

\noindent \textbf{Metrics}. We evaluate common metrics in Aerial ObjectNav~\cite{xiao2025uav,ji2025towards}: (i) \textit{Success Rate (SR)}, denoting task completion issued by the drone within a target proximity threshold; (ii) \textit{Oracle Success Rate (OSR)}, indicating whether the target threshold was achieved at least once during the mission; (iii) \textit{Success weighted by Path Length (SPL)}, capturing path optimality relative to the ground-truth geodesic distance; (iv) \textit{Distance to Goal (DTG)}, measuring residual Euclidean error upon episode termination; and (v) \textit{Flight Time (FT)} measures wall-clock time from mission start to termination including model inference latency.

\noindent \textbf{Baselines}. We compare our method against representative state-of-the-art ObjectNav baselines, using official repositories or our own re-implementations. All methods start at the same location and are provided with the same user query. Importantly, some methods have advantages such as additional sensors, access to more powerful cloud-hosted VLMs, and multiple dedicated perception models.
\textbf{CLIP-H} \cite{xiao2025uav} uses a CLIP model to compute cosine similarity between embedded multi-view drone images and a textual target description, then selects an action towards the most similar image (forward, left, right, descend). It stops when similarity exceeds a preset threshold. 
\textbf{See-Point-Fly} (SPF) \cite{hu2025see} uses a cloud-hosted VLM to predict 2D pixel coordinates of target objects on images and project them to 3D waypoints using depth estimation. We use Gemini-2.5-flash as the cloud model. 
\textbf{VLFM} \cite{yokoyama2024vlfm} builds occupancy maps from depth to detect exploration frontiers and assigns a language-aligned relevance score for the image using a VLM to guide the frontier selection. Then, they use several object detectors and segmentation models for detecting the target.

\begin{table}[t]
\caption{\textbf{Benchmark comparison}. Best results in \colorbox{green!15}{green} and second-best in \colorbox{cyan!15}{cyan}.}
\centering
\vspace{-10pt}
\renewcommand{\arraystretch}{1.2}
\resizebox{\columnwidth}{!}{%
\begin{tabular}{lcccccc}
\toprule
\textbf{Method} & \textbf{SR $\uparrow$} & \textbf{OSR $\uparrow$} & \textbf{SPL $\uparrow$} & \textbf{DTG $\downarrow$} & \textbf{FT $\downarrow$} \\
\midrule
CLIP-H \cite{xiao2025uav} & 5.6 & 6.5 & 4.9 & 42.7 & 370.5 \\
VLFM \cite{yokoyama2024vlfm} & 14.0 & 24.3 & 9.6 & \cellcolor{cyan!15}{39.1} & \cellcolor{cyan!15}{316.1} \\
SPF \cite{hu2025see} & \cellcolor{cyan!15}{22.5} & \cellcolor{cyan!15}{31.0} & \cellcolor{cyan!15}{20.7} & 48.7 & 332.6 \\
SkySearcher (ours) & \cellcolor{green!15}{\textbf{81.3}} & \cellcolor{green!15}{\textbf{81.3}} & \cellcolor{green!15}{\textbf{52.8}} & \cellcolor{green!15}{\textbf{24.7}} & \cellcolor{green!15}{\textbf{245.5}} \\
\bottomrule
\end{tabular}
}
\label{tab:overall}
\end{table}

\begin{figure}[t]
    \centering
    \includegraphics[width=\linewidth]{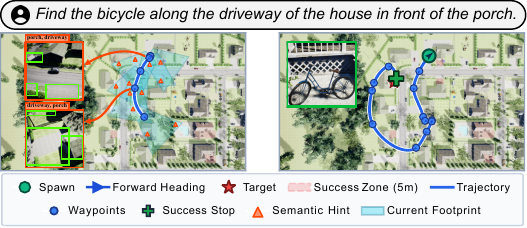}
    \vspace{-20pt}
    \caption{Qualitative results obtained with \sysname in the Neighborhood environment. Left: intermediate mission state and semantic hints (house driveway, porch). Right terminal mission state, in which the drone reaches the ground-truth target location within the prescribed accuracy.}
    \label{fig:qualitative}
\end{figure}

\begin{table}[t]
\caption{\textbf{Per-environment results.}}
\centering
\vspace{-7pt}
\renewcommand{\arraystretch}{1.2}
\resizebox{\columnwidth}{!}{%
\begin{tabular}{lcccccc}
\toprule
\textbf{Method}
& \textbf{SR $\uparrow$}
& \textbf{OSR $\uparrow$}
& \textbf{SPL $\uparrow$}
& \textbf{DTG $\downarrow$}
& \textbf{FT $\downarrow$} \\
\midrule
Cabinlake    & 84.2 & 84.2 & \textbf{66.1} & 27.2 & 271.6 \\ 
Downtown     & \textbf{86.1} & \textbf{86.1} & 56.4 & \textbf{16.0} & 226.3 \\
Neighborhood & 85.2 & 85.2 & 45.1 & 18.8  & 271.1 \\
Venice       & 68.0 & 68.0 & 45.8 & 44.8 & \textbf{225.6} \\
\bottomrule
\end{tabular}%
}
\vspace{-5pt}
\label{tab:single_envs}
\end{table}

\subsection{Benchmark Results}\label{VI-B}
Benchmark results and comparison are reported in Table~\ref{tab:overall}. \sysname outperforms all baselines significantly\footnote{Our results are also comparable to other methods with similar evaluations~\cite{chen2026airhunt}. However, we were unable to include them since no code implementation nor benchmark details are available at submission time.}, raising SR from 22.5\% (SPF, the strongest baseline) to 81.3\%. Notably, OSR does not increase for our method, being indicative of our goal-directed strategy. Since we descend only during the target approach, we do not come within the target threshold as opposed to low-altitude strategies. We also improve SPL (52.8 vs. 20.7), confirming the success rate gains do not come at the cost of path efficiency. Overall, we achieve the lowest DTG, even when we end the mission at the closest location. Finally, despite the increased initial altitude gain and required descent, our search strategy also incurs the least mission flight time. CLIP-H evaluates the global semantics of images, being unable to discern targets in expansive views. VLFM was designed for indoor setups where its different semantic modules and depth sensor can extract more concrete semantic information. We attribute the performance of SPF to the lack of a persistent map. It attains the second-best SR but, once the 2D-to-3D waypoint prediction drifts, it has no way to recover, resulting in the highest DTG.

We also report per-environment results in Table~\ref{tab:single_envs}. Performance remains consistent despite environment variability and scale, showing the generality of our method.
Downtown yields the highest SR and lowest DTG, consistent with the richer semantics (streets, buildings) that urban layouts provide for grounding. Conversely, Venice is the most challenging, primarily due to narrow alleys and frequent occlusions that break line-of-sight to candidate targets. The re-observation reward $\lambda_\mathrm{reobs}$ mitigates this by incentivizing further search on previously uncertain regions. On the other hand, Cabinlake, a wide and free environment, achieves the best SPL, suggesting that once salient cues are correctly identified, the resulting trajectories are highly efficient.

\begin{figure}[t]
    \centering
    \includegraphics[width=\linewidth]{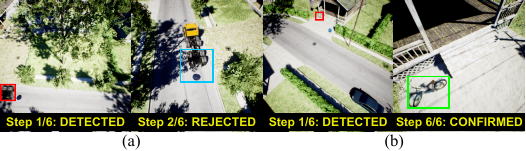}
    \vspace{-22pt}
    \caption{Example of the visual servoing module corresponding to the query "Find the bicycle along the driveway of the house in front of the porch": (a) the VLM incorrectly detects a bicycle from truck tires, but thanks to a closer verification the model recovers from a false positive; (b) the VLM correctly spots the query and navigates toward it, verifying its correctness and successfully ending the mission.}
    \label{fig:visual_servoing}
\end{figure}

We show in Fig.~\ref{fig:qualitative} a representative mission from our system. Given the instruction ``Find the bicycle along the driveway of the house in front of the porch'', \sysname explores the nearby roads, identifying driveways and house fronts as semantic cues in the environment. The drone progressively updates the target likelihood, performing a semantically guided search as candidate porches and driveways are observed and ruled out, until it localizes the bicycle and issues a stop within the success threshold. In Fig.~\ref{fig:visual_servoing} we show an example of the visual servoing process where the VLM validates target detections and recovers from outliers.

\subsection{Ablation Study}
To quantify the contribution of each term in \eqref{eq:InfGain}, we conduct an ablation study in which individual reward terms are removed from the planner's objective, while keeping the geometric term $\text{IG}_{\text{geom}}$ active in every variant to preserve basic exploration capability. We evaluate three partial configurations---\textit{Geometric}, \textit{Geometric and Reobservation}, and \textit{Geometric and Semantic}---against the \textit{Full Cost}, under the same experimental protocol of Sec.~\ref{VI-B}. Results are reported in Table~\ref{tab:ablation}.

\begin{table}[t]
\caption{\textbf{Ablation Study on IG Terms.}}
\centering
\vspace{-7pt}
\renewcommand{\arraystretch}{1.2}
\resizebox{\columnwidth}{!}{%
\begin{tabular}{lccccc}
\toprule
\textbf{Method}
& \textbf{SR $\uparrow$}
& \textbf{OSR $\uparrow$}
& \textbf{SPL $\uparrow$}
& \textbf{DTG $\downarrow$}
& \textbf{FT $\downarrow$} \\
\midrule
Geometric &  29.6 &  30.6 &  27.4 &  144.7 & 375.4 \\
Geometric and Reobservation & 53.3 & 54.2 & 35.2 & 69.3 & 331.7 \\
Geometric and Semantic &  54.2 &  56.1 &  42.0 &  70.2 & 352.0 \\
\midrule
\rowcolor{gray!20}
Full Cost  &  \textbf{81.3} &  \textbf{81.3} &  \textbf{52.8} &  \textbf{24.7} & \textbf{245.5} \\
\bottomrule
\end{tabular}%
}
\vspace{-4pt}
\label{tab:ablation}
\end{table}

With only $\text{IG}_{\text{geom}}$ active, the planner reduces to frontier-driven coverage, blind to the query: SR drops to 29.6\% and DTG rises to 144.7 m, confirming that pure geometric exploration is insufficient for semantically-grounded search. Adding either $\text{IG}_{\text{sem}}$ or $\text{IG}_{\text{reobs}}$ individually nearly doubles SR (54.2\% and 53.3\%, respectively), showing that each term captures complementary information missing from coverage alone; Semantic contributes marginally more in isolation, consistent with its role in directly exploiting the belief map rather than only resolving ambiguity in it. Notably, neither partial configuration approaches the localization precision of the full system.
This indicates that the three terms act synergistically rather than additively---geometric coverage alone cannot localize the target, semantic guidance alone cannot resolve occlusions or borderline detections, and re-observation alone has nothing to re-examine without semantic guidance directing it toward promising regions. 
Only their combination in Full Cost yields both high success (81.3\%) and efficient, low-latency missions (FT of 245.5 s, the lowest across all variants), confirming that all three reward terms are necessary and mutually reinforcing.

\subsection{Real-world deployment}

We deploy \sysname on a drone to validate that our system can run onboard a real robot platform. The drone is a quadrotor Holybro X500 frame with a Pixhawk4 flight controller. It is equipped with a RealSense D435i RGB-D camera pitched at \(-60^\circ\). As companion computer, it has an Nvidia Jetson NX with 16Gb unified RAM, being able to run Qwen3-VL-4B VLM, our map, and the planner. This setup replicates our simulated sensor and model. To simplify localization and navigation and comply with flight regulations, the drone is positioned using an RTK-GPS ground station and controlled by a human pilot towards the viewpoints suggested by our planner. No intervention or human decision is involved in the perception and planning.

\begin{figure}
    \centering
    \includegraphics[width=0.95\linewidth]{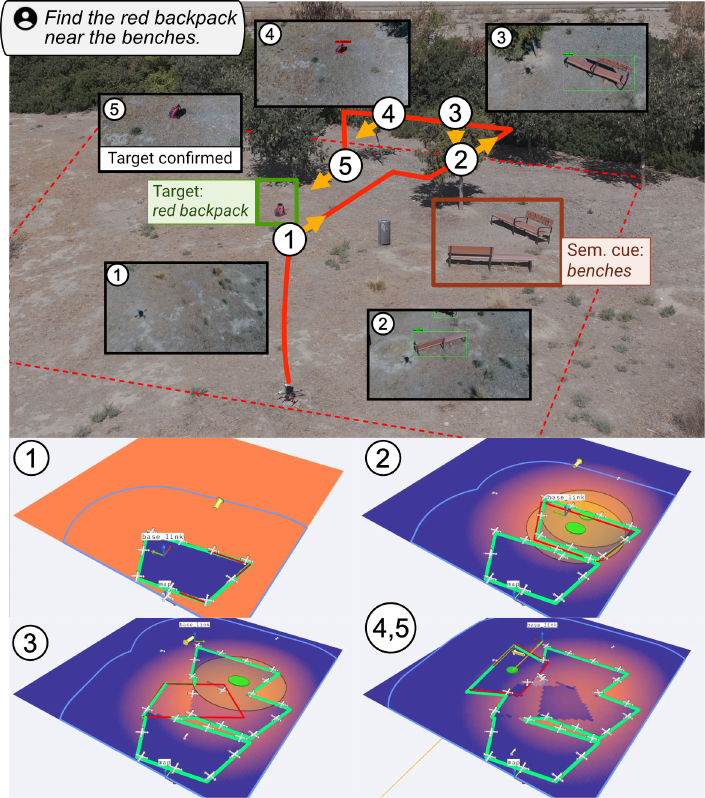}
    \vspace{-8pt}
    \caption{Real deployment of \sysname on a drone with representative viewpoints and corresponding semantic likelihood maps. The dashed red line indicates the search area, grey arrows are viewpoint candidates, yellow arrows are the NBV, green blobs are semantic hints, orange and blue means high and low target likelihood in the maps. The drone takes off and begins exploring geometrically (1) until finding a semantic hint (2). Then, it prioritizes exploring the region around the cue (3) until finding the target and performing a final approach (4,5).}
    \vspace{-4pt}
    \label{fig:real}
\end{figure}

We carry out the search mission shown in Fig.~\ref{fig:real}. The environment measures $22 \times 20$ m. The target was measured $9.40$ m north and $7.25$ m west from the starting position of the drone. The flight altitude was set to $5$ m to avoid the drone detecting the target at first glance and requiring exploration of the environment. The search begins with geometric exploration (1), after detecting a hint, the likelihood map is updated (2) and the search is directed towards semantically relevant regions (3). The mission ends with the location of the target (4) and final approach (5), resulting in a final target position at $9.25$ m north and $7$ m west from the initial location with an error below one meter.

\section{Conclusion}
This paper introduced \sysname, a high-altitude aerial system for open-vocabulary semantic search in large-scale outdoor environments.
Operating at high altitude, our solution exploits the wide field of view available from aerial platforms to reason efficiently how to explore these environments.
This is possible thanks to the combination of a small-scale Vision-Language Model (VLM) with a lightweight map representation and an information-theoretic next-best-view planner, also enabling semantic search without relying heavily on cloud infrastructure.
Experiments in high-fidelity simulation and a real-world deployment on a custom quadrotor demonstrate that \sysname consistently outperforms existing baselines in outdoor semantic search tasks.

\bibliographystyle{IEEEtran}
\bibliography{bibliography.bib}

\vfill
\end{document}